\pdfoutput=1

\documentclass[11pt]{article}

\usepackage{ACL2023}
\usepackage{multirow} 
\usepackage{times}
\usepackage{latexsym}
\usepackage[T1]{fontenc}

\usepackage[utf8]{inputenc}
\usepackage[framemethod=TikZ]{mdframed}
\usepackage{booktabs}

\usepackage{microtype}
\usepackage{longtable}
\usepackage{inconsolata}
\usepackage{amsmath}
\newcommand{\vpara}[1]{\vspace{0.2cm}\noindent\textbf{#1 }}

\title{AutoRE: Document-Level Relation Extraction \\ with Large Language Models}

\author{Lilong Xue$^{\star}$ \\
  Tsinghua University \\
  \texttt{xll21@mails.tsinghua.edu.cn} \\\And
  Dan Zhang$^{\star}$ \\
  Tsinghua University \\
  \texttt{zd21@mails.tsinghua.edu.cn} \\\AND 
  Yuxiao Dong \\
  Tsinghua University \\
  \texttt{yuxiaod@tsinghua.edu.cn}\\\And
  Jie Tang$^{\dagger}$ \\
  Tsinghua University \\
  \texttt{jietang@tsinghua.edu.cn} \\
}

\begin{document}
\maketitle
\begin{abstract}
Large Language Models (LLMs) have demonstrated exceptional abilities in comprehending and generating text, motivating numerous researchers to utilize them for Information Extraction (IE) purposes, including Relation Extraction (RE). Nonetheless, most existing methods are predominantly designed for Sentence-level Relation Extraction (SentRE) tasks, which typically encompass a restricted set of relations and triplet facts within a single sentence. Furthermore, certain approaches resort to treating relations as candidate choices integrated into prompt templates, leading to inefficient processing and suboptimal performance when tackling Document-Level Relation Extraction (DocRE) tasks, which entail handling multiple relations and triplet facts distributed across a given document, posing distinct challenges. To overcome these limitations, we introduce AutoRE, an end-to-end DocRE model that adopts a novel RE extraction paradigm named RHF (Relation-Head-Facts). Unlike existing approaches, AutoRE does not rely on the assumption of known relation options, making it more reflective of real-world scenarios. Additionally, we have developed an easily extensible RE framework using a Parameters Efficient Fine Tuning (PEFT) algorithm (QLoRA). Our experiments on the RE-DocRED dataset showcase AutoRE's best performance, achieving state-of-the-art results, surpassing TAG by 10.03\% and 9.03\% respectively on the dev and test set. 
The code is available\footnote{https://github.com/THUDM/AutoRE} and the demonstration video is provided\footnote{https://www.youtube.com/watch?v=IhKRsZUAxKk}.
\end{abstract}

{\let\thefootnote\relax\footnotetext{$^\star$LX and DZ contributed equally.}}
{\let\thefootnote\relax\footnotetext{$^\dagger$JT is the corresponding author.}}
\section{Introduction}
The rise of LLMs, such as GPT-4 \citep{Achiam2023GPT4TR} and Llama2 \citep{Touvron2023Llama2O}, has significantly propelled the progress of natural language processing due to their strong capabilities in text understanding, generation, scientific reasoning \citep{zhang2024sciglm, zhang2024rest}, social bot detection \citep{zhou2024lgb}, and generalization \citep{Zhao2023ASO}. 
There has been an increasing interest in using LLMs to generate structured information for IE tasks \citep{Xu2023LargeLM,  Wadhwa2023RevisitingRE} and making impressive progress. 
Typical IE tasks using LLMs include Named Entity Recognition (NER) \citep{wang2023gpt}, Relation Extraction (RE) \citep{zhou2023llm}, and Event Extraction (EE) \citep{Xu2023LargeLM}.
Despite the outstanding result, the performance of current LLMs in RE is still far from satisfactory.
\begin{figure}[t!]
    \centering
    \includegraphics[width=.46\textwidth]{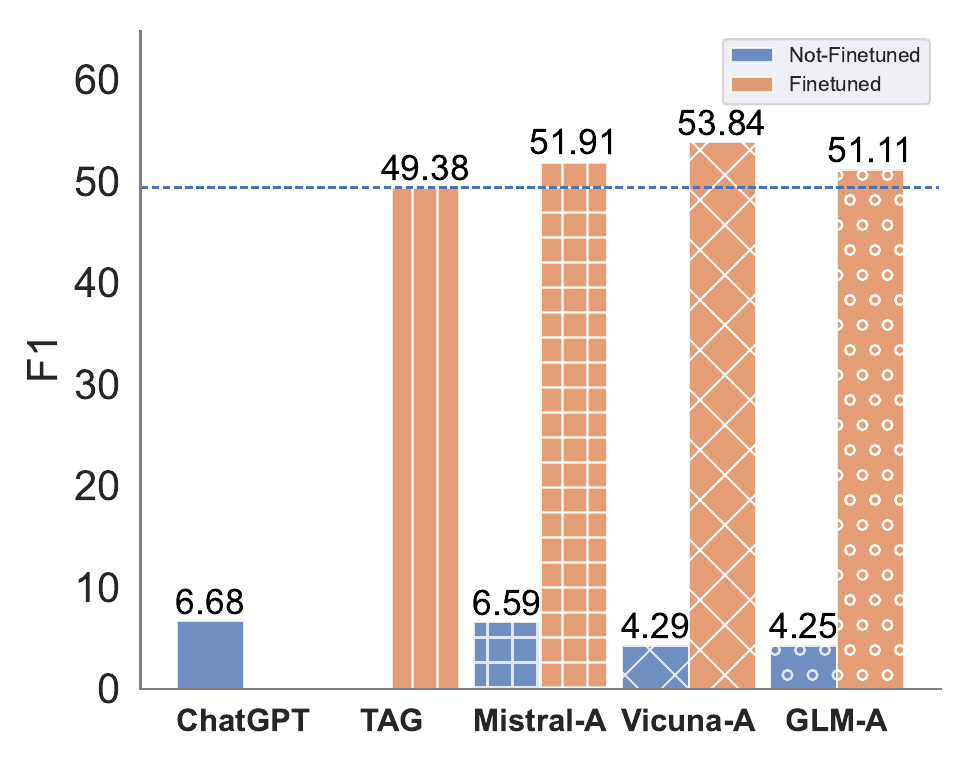}
    \caption{The result on the test set of Re-DocRED. AutoRE (-A) achieves SOTA for different LLMs.}
    \label{fig:compare}
\end{figure}

\vpara{Underperformance on DocRE Tasks.} We evaluated several high-performing LLMs on the document-level RE (DocRE) task, specifically using the test set of Re-DocRED \citep{Tan2022RevisitingD}. These models included GPT-3.5-turbo\footnote{https://chat.openai.com/chat}(ChatGPT), Mistral-7B-Instruct-v0.2 \citep{Jiang2023Mistral7} (Mistral-7B), Vicuna-7B-v1.5 \citep{vicuna2023} (Vicuna-7B), and ChatGLM3-6B \citep{du2022glm}. Our results indicate that, without specific fine-tuning, the performance of these language models on DocRE tasks is suboptimal, as shown in the blue bars in Figure \ref{fig:compare}.

\vpara{Inefficacy in Multi-Relations.} 
Incorporating relations directly into the prompt template as candidates is a common strategy for LLM-based models \citep{Wang2023InstructUIEMI, Wei2023ZeroShotIE, Xiao2023YAYIUIEAC}. This method is effective for tasks that involve a relatively small number of relation types.
However, the number of relation types can easily exceed 100 in real-world scenarios. 
Dealing with multiple relations, as seen in the Re-DocRED dataset with its 96 relation types, presents a significant challenge for most existing models. 
Embedding such many relations directly into the prompt template is often impractical \citep{Wadhwa2023RevisitingRE}.

\vpara{Limitations of Current RE Paradigms.} The current paradigms in RE exhibit significant limitations in their effectiveness. Modern generative methods typically operate by either directly producing triplet facts from the input text in a singular step \citep{Wang2023InstructUIEMI}, or by initially identifying a set of relations and subsequently generating triplet facts based on these relations \citep{Wei2023ZeroShotIE}. Earlier approaches prioritized the extraction of the head entity before the derivation of triplet facts \citep{Li2019EntityRelationEA}. 
However, these methodologies fall short of handling DocRE tasks that involve multiple relations and plenty of triplet facts. For instance, a single instance in the Re-DocRED dataset might encompass as many as 27 different relations and include up to 142 distinct triplet facts.

To address challenges identified in existing RE paradigms, we innovate a new pipeline RE paradigm, Relation-Head-Facts (RHF).
We comprehensively redefined the 96 relation descriptions and crafted simplified relation extraction templates, developing an instruction-tuning dataset based on Re-DocRED. Utilizing the Mistral-7B model with Parameter Efficient Fine-Tuning (PEFT), QLoRA \cite{Dettmers2023QLoRAEF}, our model achieved state-of-the-art (SOTA) performance on the Re-DocRED test set. 
Key contributions of our work include:

\vpara{Various RE Paradigms.}  We conducted experiments across a variety of RE paradigms and revealed that a pipeline RE approach is especially potent for DocRE, particularly RHF. This paradigm prioritizes the extraction of relations, followed by the identification of subjects, thereby significantly enhancing the model's capacity to efficiently and accurately uncover triplet facts. 

\vpara{Efficient DocRE Model.} Adopting the RHF paradigm for DocRE and refined relation descriptions, we have meticulously crafted an instruction-finetuning dataset based on Re-DocRED. This dataset was utilized to fine-tune the Mistral-7B with QLoRA, culminating in the creation of AutoRE, which achieved SOTA results across multiple pre-trained LLMs (PLMs), demonstrating the generality and effectiveness of this model architecture.


\vpara{Easy Enhancement of Capabilities.} We have 
incorporates three distinct QLoRA modules within the RHF framework, where each module is exclusively responsible for a specific task: one for relation extraction, another for head entity identification, and the third for triple fact extraction, ensuring specialized handling for each aspect. 
This strategy effectively lays the groundwork for future advancements while ensuring a minimal rise in computational demands and avoiding interference between subtasks.
\section{Related work}
DocRE refers to the task of extracting relations between entities at the document level, we follow the definition in \citep{Zheng2023ASO}:
Given a document $\mathcal{D}$  with a set of sentences containing a set of entities  $\mathcal{V}=\left\{e_{i}\right\}_{i=1}^{N}$. The DocRE task is to predict the relation types between an entity pair $\left(e_{h}, e_{t}\right)_{h, t \in\{1, \cdots, N\}, h \neq t} $, where $h$ stands for the head (subject) and $t$ stands for the tail (object). 

\vpara{LLMs for DocRE.} Researchers have been employing LLMs to tackle RE tasks. For example, ChatIE \citep{Wei2023ZeroShotIE} deconstructs the complex RE process into assembling the outputs from multiple rounds of Question-Answer into a final structured format.
PromptRE \citep{gao2023promptre} integrates LLM prompting with data programming to deal with DocRE. However, the performance of LLMs on RE tasks still lags behind SOTA models. \citeauthor{Han2023IsIE} concluded that \textit{ChatGPT does not adequately comprehend the subject-object relationships in RE tasks}. Similarly, \citeauthor{Li2023EvaluatingCI} noted that in Standard-IE settings, \textit{ChatGPT's performance is generally not as effective as BERT-based models}. Moreover, most models are tested on SentRE. To test LLMs for DocRE, we conducted tests on ChatGPT, Mistral-7B, Vicuna-7B, and ChatGLM3-6B and revealed that the current performance is far from satisfactory, as illustrated in Figure \ref{fig:compare}. This aligns with findings reported by \citep{Li2023SemiautomaticDE}, indicating that current models still fall significantly short in performance on DocRE.


\vpara{RE Prompt Template.} These models fine-tuned on LLMs for RE operate on a prompt-based or instruction-driven mechanism \citep{beurer2023prompting}, engaging in a question-and-answer format to execute RE tasks.
ChatIE \citep{Wei2023ZeroShotIE}, InstructUIE \citep{Wang2023InstructUIEMI}, and YAYI \cite{Xiao2023YAYIUIEAC} while demonstrating formidable capabilities in IE, exhibit considerable limitations in their RE prompt templates. A common method in their RE process involves embedding a list of relations into the model's prompt template as alternatives. However, this approach becomes impractical when dealing with DocRE, such as the 96 relations in the Re-DocRED. This limitation is acknowledged by \citeauthor{Wadhwa2023RevisitingRE}, who concludes that \textit{``for datasets with long texts or a large number of targets, it is not possible to fit detailed instructions in the prompt''}.

\vpara{RE Paradigms.}
Within the context of LLMs, RE paradigms are primarily categorized into two types: Pipeline and Joint. The Pipeline approach involves first identifying relations and then extracting triplet facts, or initially extracting a head entity followed by its corresponding relation and tail entity. This approach deviates from the traditional methodology of first extracting entities and then determining their interrelations \citep{chen2022pattern,jiang2020targeting}. The main drawbacks is that applying the conventional Pipeline approach to LLMs can be extremely time-consuming, particularly when many entities lack interrelations. On the other hand, the Joint paradigm, which inputs a text and directly outputs all triplet facts as seen in \citep{Zhang2023ANT}, aligns more closely with traditional practices. However, as illustrated in Table \ref{tab:chatgpt-result}, these paradigms encounter significant challenges when applied to DocRE, particularly due to the complexity of handling samples that may contain multiple relations and a multitude of triplet facts.

In summary, current LLMs still exhibit significant gaps in performance for DocRE, indicating a need for further fine-tuning. Additionally, the existing RE templates, which treat relations as candidates, struggle to handle scenarios involving multiple relations. Coupled with the underwhelming effectiveness of current RE paradigms, there is a need for a paradigm shift. 

\section{Methodology} 

\subsection{RE Paradigms} 
\label{rep}
\begin{figure*}[t]
    \centering
    \includegraphics[width=.7\textwidth]{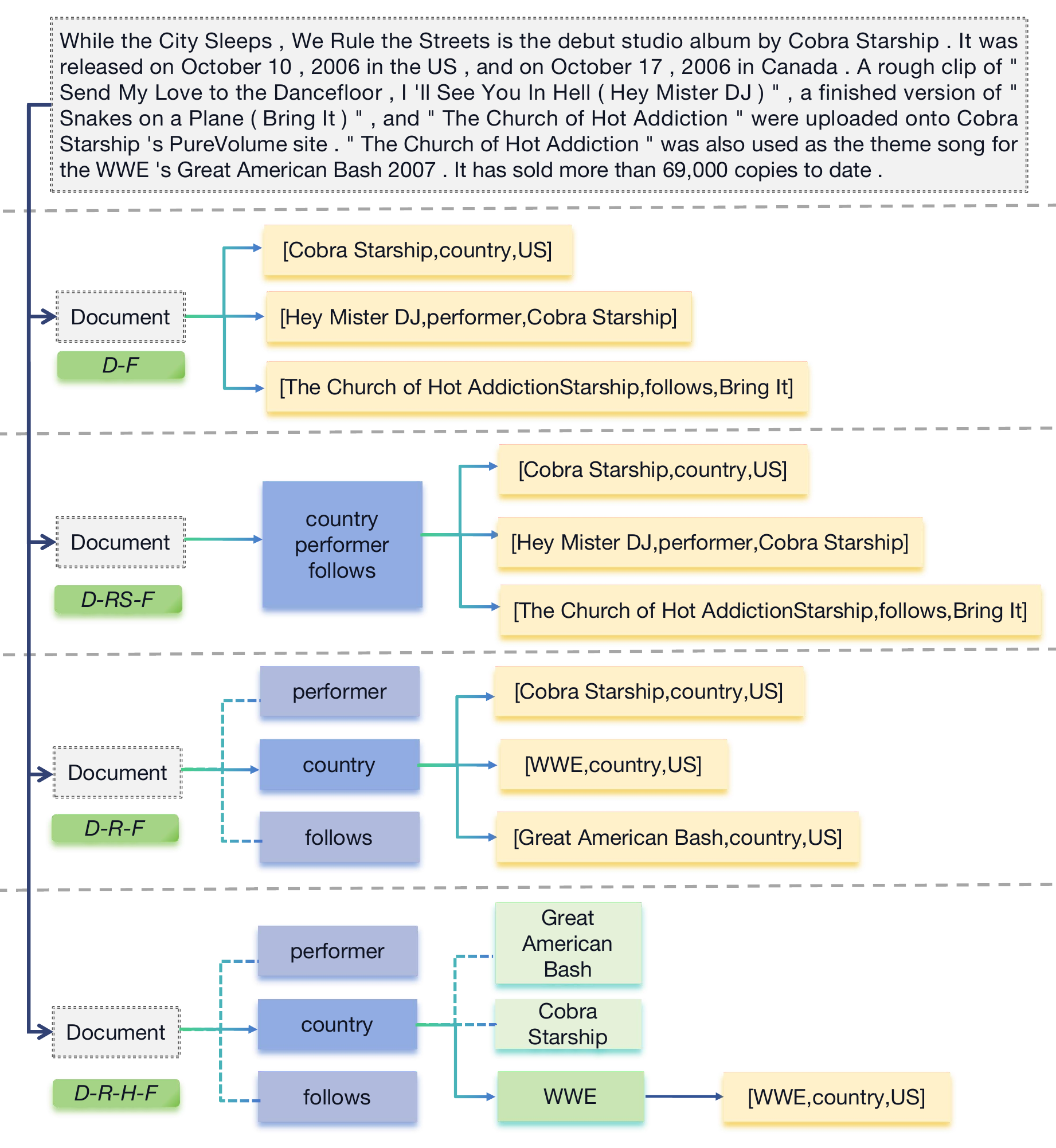}
    \caption{Processing steps of different RE paradigms.}
    \label{fig:RE_Paradigms}
\end{figure*}

We summarized the existing paradigms of RE and designed a unique extraction paradigm, different RE paradigms are illustrated in Figure \ref{fig:RE_Paradigms}. 

\vpara{Document-facts (D-F).} Fed with a document, the model directly outputs all triplets facts. This method is brute-force and requires the shortest inference time. It directly inputs relation types as candidates into the prompt and then let the model generate all triplet facts in one step as InstructIE \citep{Wang2023InstructUIEMI} did. 

\vpara{Document-relations-facts (D-RS-F).} In this paradigm, the model extracts the relations present within the document and embeds all the predicted relations into the prompt to obtain triplet facts.

\vpara{Document-relation-facts (D-R-F).} In this framework, the model identifies relations within a given sentence and systematically traverses these relations to acquire triplet facts that correspond to each identified relation, which is similar to the approach taken by \cite{Wei2023ZeroShotIE}.

\vpara{Document-relation-head-facts (D-R-H-F).} In our newly designed paradigm, the model specifically focuses on each relation to identify an appropriate set of entities that will function as the \verb|'head'| in the triplet facts. Subsequently, the relevant triplets facts corresponding to these relations are extracted.

\begin{table}
\centering
\resizebox{!}{13mm}{
\begin{tabular}{lccccc}
\toprule
\textbf{Paradigm} & \textbf{TP} & \textbf{FP} & \textbf{R} & \textbf{P} & \textbf{F1}\\
\toprule
D-F & 735 & 3824 & 4.21 & 16.12 & 6.68\\
D-RS-F & 867 & 4811 & 4.97 & 15.27 & 7.50 \\
D-R-F & 1674 & 93741 & 9.59 & 1.75 & 2.97 \\
D-R-H-F & 3201 & 333226 & 18.35 & 0.95 & 1.81 \\
\bottomrule
\end{tabular}
}
\caption{The result of four RE paradigms with ChatGPT. 
Here, TP denotes True Positive, FP is False Positive, R for Recall, P means Precision, and F1 references Micro F1. All paradigms perform poorly.  }
\label{tab:chatgpt-result}
\end{table}


We test these paradigms with ChatGPT and the results are displayed in Table \ref{tab:chatgpt-result}. We provide testing prompts for the Re-DocRED dataset under different paradigms using ChatGPT in Table \ref{tab:chat-template}. For brevity, we only provide two representative relation extraction prompt words. The rest are similar to these.
We arrived at the following conclusion: 
\begin{itemize}
    \item LLMs still perform poorly in DocRE tasks involving the extraction of multiple relations and triplet facts, achieving only single-digit scores. As of now, fine-tuning the model is still necessary.
    \item By extending the thought chain to derive final triplet facts, we can obtain more accurate triplet facts, though this approach does introduce a higher number of erroneous facts. 
    \item Harnessing the last paradigm, which we name RHF, the model can find triplets facts more accurately in a step-by-step mode with finer-grained tasks, thereby enhancing recall rates. 
\end{itemize}

\subsection{Dataset Processing} We used the Re-DocRED dataset for fine-tuning, refining it by removing duplicates and ensuring factual accuracy. This involved adjusting reciprocal relations like ``follows'' and ``followed by'' to accurately represent inversion, enhancing the dataset's robustness and precision.

In earlier experiments with ChatGPT, we discovered that providing the model with descriptions of relations enhances its capability to extract factual information. Nevertheless, incorporating Wikidata relation descriptions\footnote{https://www.wikidata.org/} led to diminished performance, likely due to their occasional lack of clarity and precision, as exemplified by:

    \textit{``located in the administrative territorial entity'': ``The item is located on the territory of the following administrative entity. Use P276 for specifying locations that are non-administrative places and for items about events. Use P1382 if the item falls only partially into the administrative entity.''}

Addressing this, we systematically rewrote all 96 relation descriptions, markedly improving model performance in Table \ref{tab:desc-result}. 
An example of our revised description is as follows. 
Details of all relation descriptions are presented in Table \ref{tab:Relation Description}.

\textit{``located in the administrative territorial entity'': ``This relation indicates that a subject (e.g., a place, event, or item) is situated within an administrative region, the object.
Example: (Harvard University, located in the administrative territorial entity, Cambridge, Massachusetts).''}

Finally, in line with the RHF paradigm, we crafted instruction fine-tuning templates, breaking down the RE process of each sample into three distinct steps. The specific details of these templates can be found in 
Table \ref{tab:instructtuning-template}, we provide a display of how we constructed our training data using simple prompt templates, the extraction of relations, the extraction of the head entity, and finally the triplet extraction.

\begin{table}
\centering
\resizebox{!}{16mm}{
\begin{tabular}{lccccc}
\toprule
\textbf{Paradigm} & \textbf{TP} & \textbf{FP} & \textbf{R} & \textbf{P} & \textbf{F1}\\
\toprule
D-R-F-no$_\text{desc}$ &1952	&27584	&11.19 	&6.61 	&8.31  \\
D-R-H-F-no$_\text{desc}$ &4005	&123631	&22.95 	&3.14 	&5.52  \\
\midrule
D-R-F-\text{wiki}$_\text{desc}$ &1296	&21482	&7.43 	&5.69 	&6.44 $\downarrow$  \\
D-R-H-F-\text{wiki}$_\text{desc}$ & 3283	&137462	&18.82 	&2.33 	&4.15$\downarrow$  \\
\midrule
D-R-F-\text{new}$_\text{desc}$ & 3508	&29002	&20.11 	&10.79 	&\textbf{14.04} $\uparrow$  \\
D-R-H-F-\text{new}$_\text{desc}$ & 4200	&118243	&24.07 	&3.43 	&\textbf{6.00} $\uparrow$  \\
\bottomrule
\end{tabular}
}
\caption{The result of two RE paradigms, we skip the step of extracting relations and instead use the correct relation as prior knowledge.}
\label{tab:desc-result}
\end{table}

\begin{table}[t!]
\centering
\resizebox{!}{18mm}{
\begin{tabular}{lcccccc}
\toprule
\textbf{Module}  & \textbf{TP} & \textbf{FP} & \textbf{Recall} & \textbf{Precision} & \textbf{F1}\\
\toprule
QLoRA-relation-dev  & 3190 & 657 & 63.81 & 82.92 & 72.12\\
QLoRA-head-dev  & 11269&	1910&	65.38& 	85.51 &	74.10  \\
QLoRA-fact-dev  & 14439&	2628	&83.77 	&84.60 &	84.18  \\
\midrule
QLoRA-relation-test  & 3073 & 686 & 64.44 & 81.75 & 72.06\\
QLoRA-head-test  & 12820 &2771 & 73.48 & 82.23 & 77.60 \\
QLoRA-fact-test  & 14439 &2628 & 82.75 & 84.60 & 83.66 \\
\midrule
AutoRE-dev  & 7588&	3805	&44.02& 	66.60 &	53.01  \\
AutoRE-test  & 7445 &	3794&	42.67 &	66.24 	&51.91  \\
\bottomrule
\end{tabular}
}
\caption{The results of AutoRE on the Re-DocRED dev and test sets for the three subtasks of RHF.}
\label{tab:autore-result}
\end{table}

\subsection{QLoRA Tuning}
Mistral-7B was selected as the foundation for fine-tuning because it demonstrated the best performance among the several open-source models tested when evaluating LLMs on the Re-DocRED task.
To facilitate efficient training, we opted for PEFT's QLoRA. The key advantage of QLoRA is its ability to combine the benefits of quantization and Low-Rank Adaptation \citep{Hu2021LoRALA}, resulting in efficient fine-tuning. Specifically, quantization reduces data complexity, allowing for more efficient storage and processing, which is particularly valuable for deploying large models on resource-constrained devices.

We leveraged three distinct QLoRA modules, each tailored to a specific stage of the RHF steps. This implementation was critical in enhancing RE efficiency. With the data volume varying across the intertwined tasks, a one-size-fits-all approach could have compromised performance. However, the modular structure of QLoRA facilitated smooth integration with the underlying base model. As a result, we instituted three distinct QLoRA modules, each meticulously fine-tuned to its specific dataset. This meticulous approach resulted in the creation of the AutoRE, which amalgamates these modules for amplified DocRE performance.
\section{Experiment} 
\subsection{Experimental Setup} 
\vpara{Test set.} In our evaluation, we utilized the refined Re-DocRED test set consisting of 499 articles and 17,448 triplet facts, and a validation set encompassing 498 articles with 17,236 triplets, ensuring a comprehensive and precise assessment.

\vpara{Evaluation Metric.} We adopted the strict Micro F1 criterion, recognizing a prediction as correct only if it precisely captures the entire relation, along with both the head and tail entities. It's important to highlight that within the Re-DocRED dataset, a triplet fact may encompass multiple aliases (mentions) for both the head and tail entities. Consequently, our evaluation protocol deems a prediction accurate if it correctly identifies any valid triplet pair. If a prediction aligns with any alias pair of the head and tail entity, it's counted as correct, but alternate accurate aliases aren't tallied in the correct statistics. Conversely, all incorrect predictions are flagged as false positives. This method ensures a stringent and statistically valid evaluation, lending robust credibility to the final results.

\begin{table}[t!]
\centering
\begin{tabular}{lcc}
\hline
\textbf{Model}  & \textbf{dev F1} & \textbf{test F1}\\
\hline
\verb|TAG| & 49.34 & 49.38 \\
\verb|AutoRE-ChatGLM3-6B| & \textbf{49.86} & \textbf{51.11} \\
\verb|AutoRE-Mistral-7B| & \textbf{53.01} & \textbf{51.91} \\
\verb|AutoRE-Vicuna-7B| & \textbf{54.29} & \textbf{53.84} \\
\hline
\end{tabular}
\begin{tabular}{lcccccc}
\hline
\end{tabular}
\caption{The results of AutoRE for different PLMs. Compared with TAG, all AutoRE models achieve the best performance. }
\label{tab:tag}
\end{table}

\begin{figure*}[t!]
    \centering
\includegraphics[width=1.0\textwidth]{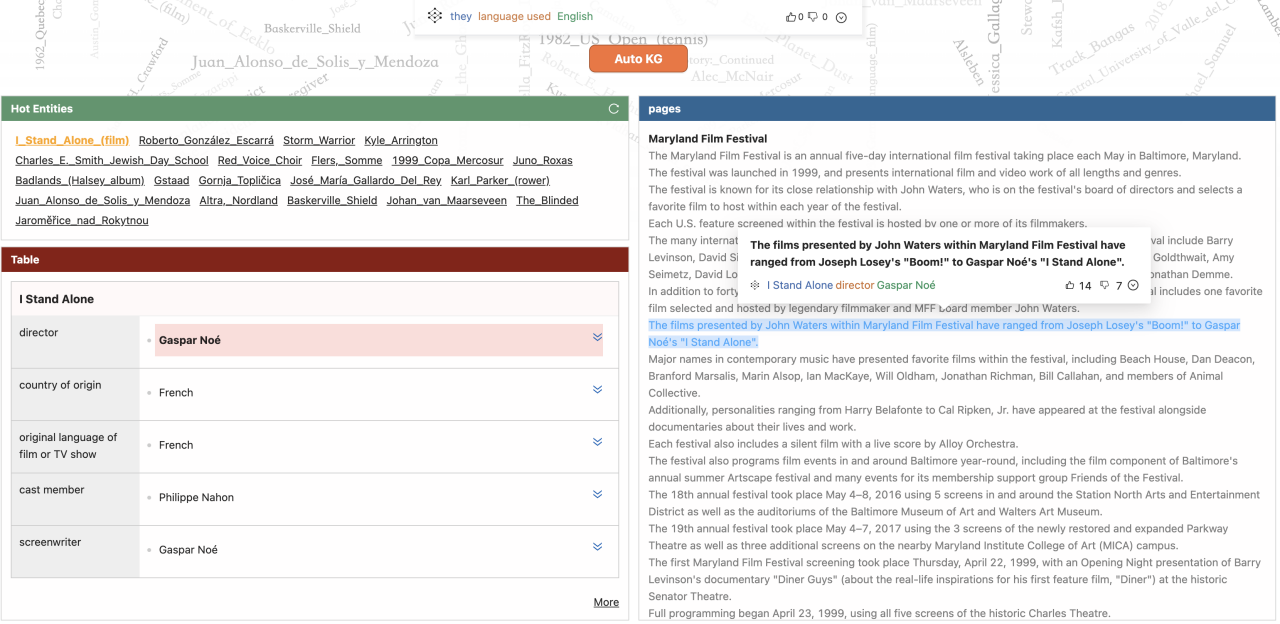}
    \caption{The homepage of online AutoRE.}
    \label{fig:online}
\end{figure*}

\begin{figure*}[t!]
    \centering
    \includegraphics[width=.85\textwidth]{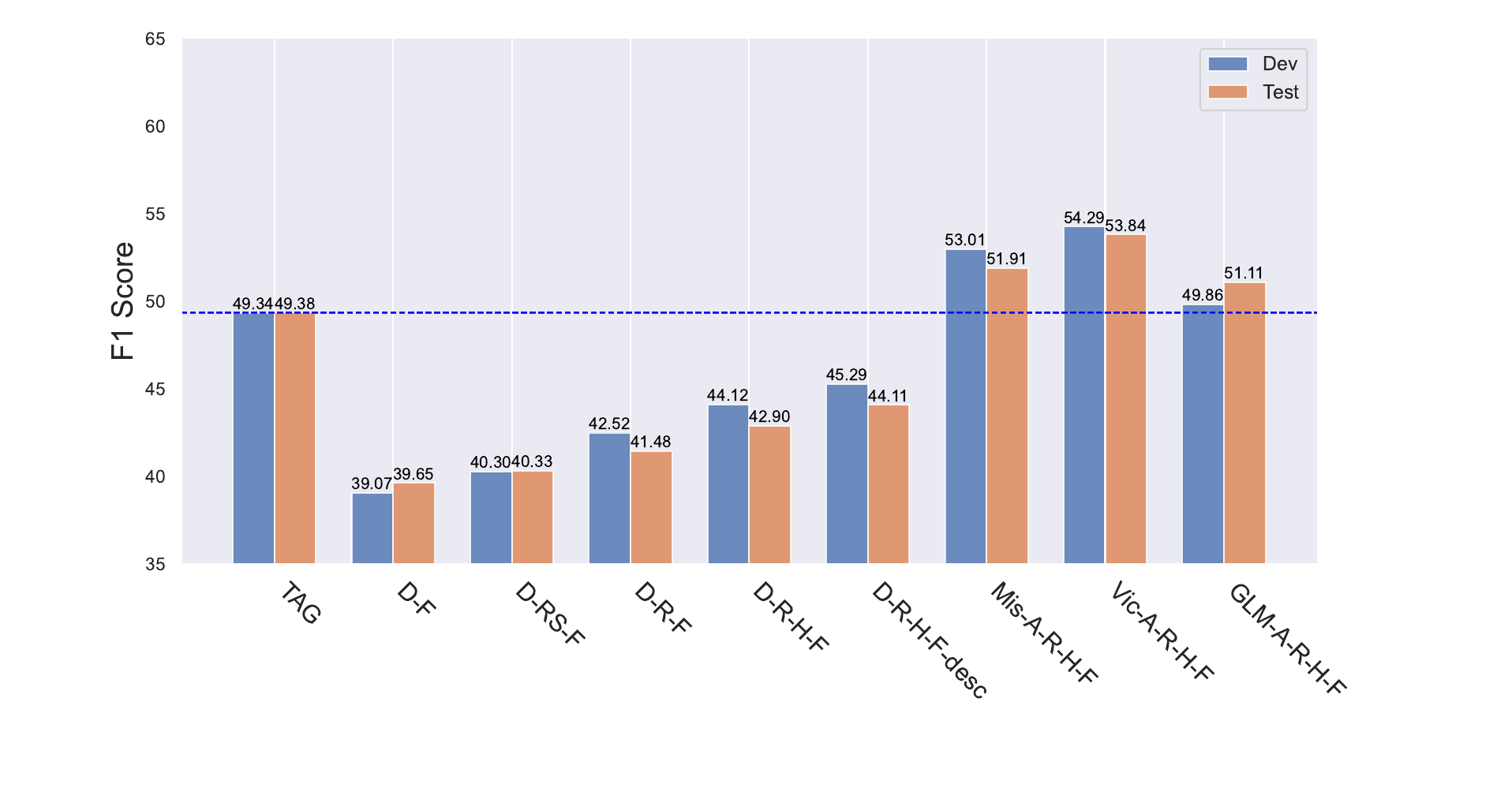}
    \caption{Performance of different paradigms and AutoRE (-A) for different PLMs. }
    \label{fig:Ablations}
\end{figure*}

\subsection{Overall Result} After training three distinct QLoRA modules, we test the performance on the Re-DocRED and then combine three QLoRAs to get the final performance on the dev set and test set, the result is shown in Table \ref{tab:autore-result}. 
When compared with TAG \citep{Zhang2023ANT} as a baseline which firstly reported the end-to-end RE on Re-DocRED, our method has achieved SOTA results, as shown in Table \ref{tab:tag}. In both the dev set and test set, the performance improvement of AutoRE finetuned with Mistral-7B over TAG is approximately 7.44\% on the dev set, and about 5.12\% on the test set demonstrating the effectiveness of our approach. Furthermore, by decomposing the task into three subtasks and training with three LoRA modules, we not only achieved excellent results but also naturally acquired an easily extendable trait. This allows for targeted improvement of a specific module's performance without impacting the performance of other subtasks. 
Additionally, it is worth noting that our work is the first to utilize large language models for processing the Re-DocRED dataset.  
AutoRE can serve as a reference for subsequent research in this field.

\subsection{Ablation}
In the ablation study, we employed Mistral-7B to fine-tune the paradigms mentioned before, revealing that the RHF model yields the best performance when solely utilizing one QLoRA module. This finding substantiates our initial hypothesis during paradigm selection: employing a step-by-step approach enhances the extraction of triplet facts while significantly reducing erroneous triplets through fine-tuning. Building on this, we compared the impact of including descriptions versus omitting them. The results confirmed that incorporating proper relation descriptions indeed benefits the model, as shown in Figure \ref{fig:Ablations}. Additionally, we explored the effectiveness of training the entire dataset with one QLoRA versus independently training different stages of RHF with three distinct QLoRAs. The latter approach demonstrated superior performance. We believe this is due to the data imbalance among predicting relations, predicting head entities, and predicting factual triples in the RHF paradigm, with the data volume for the three subtasks begin approximately 2.8\%, 24.23\%, and 72.97\%, respectively.
When combined, the model tends to favor the prediction of triples, while its capability to predict relations is relatively insufficient. 

Additionally, we have applied this framework to Vicuna-7B and ChatGLM3-6B, and both models surpassed the current SOTA levels, demonstrating the universality of the AutoRE framework. The comparative results of these experiments are illustrated in the accompanying Figure \ref{fig:Ablations}. Vicuna-7B scored the highest,
surpassing TAG by 10.03\% and 9.03\% respectively on the dev and test set, whereas ChatGLM3-6B was somewhat lower. This might be due to ChatGLM3-6B having a higher proportion of Chinese in its pre-training, while it was tested on an English task. We have deployed the system on the online platform\footnote{https://models.aminer.cn/neptune/} for users to access and experience, as shown in Figure \ref{fig:online}.

\section{Conclusion}
In this paper, we introduce RHF, a new paradigm for RE, alongside AutoRE, an advanced DocRE model. AutoRE represents a cutting-edge approach to the DocRE task, utilizing LLMs combined with QLoRA. This innovative model establishes a new standard, achieving SOTA results on the Re-DocRED dataset. AutoRE proficiently addresses the intricate task of extracting multiple relations from document-level texts, a significant challenge that has stymied existing models. Our future goal is to create a comprehensive, unified framework for RE, fully leveraging the capabilities and promise of this paradigm.

\section*{Limitations}


\vpara{Insufficient Number of Relations.} In real-world applications, the number of relations can reach thousands, significantly surpassing the 96 relations present in the Re-DocRED dataset. To better adapt to these real-world scenarios, it is imperative to gather more extensive datasets and expand the range of relations.

\vpara{Limitation to In-Domain Data.} AutoRE is not equipped to handle unseen relations. This limitation underscores the method's inadequate generalizability, primarily due to the limited scope of data it has been trained on.



\section*{Acknowledgements}
This work is supported by Technology and Innovation Major Project of the Ministry of Science and Technology of China under Grant 2022ZD0118600, Natural Science Foundation of China(NSFC) 62276148 and 62425601, the New Cornerstone Science Foundation through the XPLORER PRIZE.

\bibliography{acl_main}

\begin{thebibliography}{29}
\expandafter\ifx\csname natexlab\endcsname\relax\def\natexlab#1{#1}\fi

\bibitem[{Achiam et~al.(2023)Achiam, Adler, Agarwal, Ahmad, Akkaya, Aleman, Almeida, Altenschmidt, Altman, Anadkat et~al.}]{Achiam2023GPT4TR}
Josh Achiam, Steven Adler, Sandhini Agarwal, Lama Ahmad, Ilge Akkaya, Florencia~Leoni Aleman, Diogo Almeida, Janko Altenschmidt, Sam Altman, Shyamal Anadkat, et~al. 2023.
\newblock Gpt-4 technical report.
\newblock \emph{arXiv preprint arXiv:2303.08774}.

\bibitem[{Beurer-Kellner et~al.(2023)Beurer-Kellner, Fischer, and Vechev}]{beurer2023prompting}
Luca Beurer-Kellner, Marc Fischer, and Martin Vechev. 2023.
\newblock Prompting is programming: A query language for large language models.
\newblock \emph{Proceedings of the ACM on Programming Languages}, 7(PLDI):1946--1969.

\bibitem[{Chen and Guo(2022)}]{chen2022pattern}
Zheng Chen and Changyu Guo. 2022.
\newblock A pattern-first pipeline approach for entity and relation extraction.
\newblock \emph{Neurocomputing}, 494:182--191.

\bibitem[{Chiang et~al.(2023)Chiang, Li, Lin, Sheng, Wu, Zhang, Zheng, Zhuang, Zhuang, Gonzalez, Stoica, and Xing}]{vicuna2023}
Wei-Lin Chiang, Zhuohan Li, Zi~Lin, Ying Sheng, Zhanghao Wu, Hao Zhang, Lianmin Zheng, Siyuan Zhuang, Yonghao Zhuang, Joseph~E. Gonzalez, Ion Stoica, and Eric~P. Xing. 2023.
\newblock Vicuna: An open-source chatbot impressing gpt-4 with 90\%* chatgpt quality.

\bibitem[{Dettmers et~al.(2023)Dettmers, Pagnoni, Holtzman, and Zettlemoyer}]{Dettmers2023QLoRAEF}
Tim Dettmers, Artidoro Pagnoni, Ari Holtzman, and Luke Zettlemoyer. 2023.
\newblock Qlora: Efficient finetuning of quantized llms.
\newblock \emph{ArXiv}, abs/2305.14314.

\bibitem[{Du et~al.(2022)Du, Qian, Liu, Ding, Qiu, Yang, and Tang}]{du2022glm}
Zhengxiao Du, Yujie Qian, Xiao Liu, Ming Ding, Jiezhong Qiu, Zhilin Yang, and Jie Tang. 2022.
\newblock Glm: General language model pretraining with autoregressive blank infilling.
\newblock In \emph{Proceedings of the 60th Annual Meeting of the Association for Computational Linguistics (Volume 1: Long Papers)}, pages 320--335.

\bibitem[{Gao et~al.(2023)Gao, Fan, Sun, and Wang}]{gao2023promptre}
Chufan Gao, Xulin Fan, Jimeng Sun, and Xuan Wang. 2023.
\newblock Promptre: Weakly-supervised document-level relation extraction via prompting-based data programming.
\newblock \emph{arXiv preprint arXiv:2310.09265}.

\bibitem[{Han et~al.(2023)Han, Peng, Yang, Wang, Liu, and Wan}]{Han2023IsIE}
Ridong Han, Tao Peng, Chaohao Yang, Benyou Wang, Lu~Liu, and Xiang Wan. 2023.
\newblock Is information extraction solved by chatgpt? an analysis of performance, evaluation criteria, robustness and errors.
\newblock \emph{ArXiv}, abs/2305.14450.

\bibitem[{Hu et~al.(2021)Hu, Shen, Wallis, Allen-Zhu, Li, Wang, and Chen}]{Hu2021LoRALA}
J.~Edward Hu, Yelong Shen, Phillip Wallis, Zeyuan Allen-Zhu, Yuanzhi Li, Shean Wang, and Weizhu Chen. 2021.
\newblock Lora: Low-rank adaptation of large language models.
\newblock \emph{ArXiv}, abs/2106.09685.

\bibitem[{Jiang et~al.(2023)Jiang, Sablayrolles, Mensch, Bamford, Chaplot, de~Las~Casas, Bressand, Lengyel, Lample, Saulnier, Lavaud, Lachaux, Stock, Scao, Lavril, Wang, Lacroix, and Sayed}]{Jiang2023Mistral7}
Albert~Qiaochu Jiang, Alexandre Sablayrolles, Arthur Mensch, Chris Bamford, Devendra~Singh Chaplot, Diego de~Las~Casas, Florian Bressand, Gianna Lengyel, Guillaume Lample, Lucile Saulnier, L'elio~Renard Lavaud, Marie-Anne Lachaux, Pierre Stock, Teven~Le Scao, Thibaut Lavril, Thomas Wang, Timoth{\'e}e Lacroix, and William~El Sayed. 2023.
\newblock Mistral 7b.
\newblock \emph{ArXiv}, abs/2310.06825.

\bibitem[{Jiang et~al.(2020)Jiang, D'Souza, Auer, and Downie}]{jiang2020targeting}
Ming Jiang, Jennifer D'Souza, S{\"o}ren Auer, and J~Stephen Downie. 2020.
\newblock Targeting precision: A hybrid scientific relation extraction pipeline for improved scholarly knowledge organization.
\newblock \emph{Proceedings of the Association for Information Science and Technology}, 57(1):e303.

\bibitem[{Li et~al.(2023{\natexlab{a}})Li, Fang, Yang, Wang, Ye, Zhao, and Zhang}]{Li2023EvaluatingCI}
Bo~Li, Gexiang Fang, Yang Yang, Quansen Wang, Wei Ye, Wen Zhao, and Shikun Zhang. 2023{\natexlab{a}}.
\newblock Evaluating chatgpt's information extraction capabilities: An assessment of performance, explainability, calibration, and faithfulness.
\newblock \emph{ArXiv}, abs/2304.11633.

\bibitem[{Li et~al.(2023{\natexlab{b}})Li, Jia, and Zheng}]{Li2023SemiautomaticDE}
Junpeng Li, Zixia Jia, and Zilong Zheng. 2023{\natexlab{b}}.
\newblock Semi-automatic data enhancement for document-level relation extraction with distant supervision from large language models.
\newblock In \emph{Conference on Empirical Methods in Natural Language Processing}.

\bibitem[{Li et~al.(2019)Li, Yin, Sun, Li, Yuan, Chai, Zhou, and Li}]{Li2019EntityRelationEA}
Xiaoya Li, Fan Yin, Zijun Sun, Xiayu Li, Arianna Yuan, Duo Chai, Mingxin Zhou, and Jiwei Li. 2019.
\newblock Entity-relation extraction as multi-turn question answering.
\newblock \emph{ArXiv}, abs/1905.05529.

\bibitem[{Tan et~al.(2022)Tan, Xu, Bing, Ng, and Aljunied}]{Tan2022RevisitingD}
Qingyu Tan, Lu~Xu, Lidong Bing, Hwee~Tou Ng, and Sharifah~Mahani Aljunied. 2022.
\newblock Revisiting docred - addressing the false negative problem in relation extraction.
\newblock In \emph{Conference on Empirical Methods in Natural Language Processing}.

\bibitem[{Touvron et~al.(2023)Touvron, Martin, Stone, Albert, Almahairi, Babaei, Bashlykov, Batra, Bhargava, Bhosale et~al.}]{Touvron2023Llama2O}
Hugo Touvron, Louis Martin, Kevin Stone, Peter Albert, Amjad Almahairi, Yasmine Babaei, Nikolay Bashlykov, Soumya Batra, Prajjwal Bhargava, Shruti Bhosale, et~al. 2023.
\newblock Llama 2: Open foundation and fine-tuned chat models.
\newblock \emph{arXiv preprint arXiv:2307.09288}.

\bibitem[{Wadhwa et~al.(2023)Wadhwa, Amir, and Wallace}]{Wadhwa2023RevisitingRE}
Somin Wadhwa, Silvio Amir, and Byron~C. Wallace. 2023.
\newblock Revisiting relation extraction in the era of large language models.
\newblock \emph{Proceedings of the conference. Association for Computational Linguistics. Meeting}, 2023:15566--15589.

\bibitem[{Wang et~al.(2023{\natexlab{a}})Wang, Sun, Li, Ouyang, Wu, Zhang, Li, and Wang}]{wang2023gpt}
Shuhe Wang, Xiaofei Sun, Xiaoya Li, Rongbin Ouyang, Fei Wu, Tianwei Zhang, Jiwei Li, and Guoyin Wang. 2023{\natexlab{a}}.
\newblock Gpt-ner: Named entity recognition via large language models.
\newblock \emph{arXiv preprint arXiv:2304.10428}.

\bibitem[{Wang et~al.(2023{\natexlab{b}})Wang, Zhou, Zu, Xia, Chen, Zhang, Zheng, Ye, Zhang, Gui, Kang, Yang, Li, and Du}]{Wang2023InstructUIEMI}
Xiao Wang, Wei Zhou, Can Zu, Han Xia, Tianze Chen, Yuan Zhang, Rui Zheng, Junjie Ye, Qi~Zhang, Tao Gui, Jihua Kang, J.~Yang, Siyuan Li, and Chunsai Du. 2023{\natexlab{b}}.
\newblock Instructuie: Multi-task instruction tuning for unified information extraction.
\newblock \emph{ArXiv}, abs/2304.08085.

\bibitem[{Wei et~al.(2023)Wei, Cui, Cheng, Wang, Zhang, Huang, Xie, Xu, Chen, Zhang, Jiang, and Han}]{Wei2023ZeroShotIE}
Xiang Wei, Xingyu Cui, Ning Cheng, Xiaobin Wang, Xin Zhang, Shen Huang, Pengjun Xie, Jinan Xu, Yufeng Chen, Meishan Zhang, Yong Jiang, and Wenjuan Han. 2023.
\newblock Zero-shot information extraction via chatting with chatgpt.
\newblock \emph{ArXiv}, abs/2302.10205.

\bibitem[{Xiao et~al.(2023)Xiao, Wang, Xu, Wang, Yang, Wang, Luo, Wang, Mao, and Zeng}]{Xiao2023YAYIUIEAC}
Xinglin Xiao, Yijie Wang, Nan Xu, Yuqi Wang, Hanxuan Yang, Minzheng Wang, Yin Luo, Lei Wang, Wenji Mao, and Daniel Zeng. 2023.
\newblock Yayi-uie: A chat-enhanced instruction tuning framework for universal information extraction.
\newblock \emph{ArXiv}, abs/2312.15548.

\bibitem[{Xu et~al.(2023)Xu, Chen, Peng, Zhang, Xu, Zhao, Wu, Zheng, and Chen}]{Xu2023LargeLM}
Derong Xu, Wei Chen, Wenjun Peng, Chao Zhang, Tong Xu, Xiangyu Zhao, Xian Wu, Yefeng Zheng, and Enhong Chen. 2023.
\newblock Large language models for generative information extraction: A survey.
\newblock \emph{arXiv preprint arXiv:2312.17617}.

\bibitem[{Zhang et~al.(2024{\natexlab{a}})Zhang, Hu, Zhoubian, Du, Yang, Wang, Yue, Dong, and Tang}]{zhang2024sciglm}
Dan Zhang, Ziniu Hu, Sining Zhoubian, Zhengxiao Du, Kaiyu Yang, Zihan Wang, Yisong Yue, Yuxiao Dong, and Jie Tang. 2024{\natexlab{a}}.
\newblock Sciglm: Training scientific language models with self-reflective instruction annotation and tuning.
\newblock \emph{arXiv preprint arXiv:2401.07950}.

\bibitem[{Zhang et~al.(2024{\natexlab{b}})Zhang, Zhoubian, Yue, Dong, and Tang}]{zhang2024rest}
Dan Zhang, Sining Zhoubian, Yisong Yue, Yuxiao Dong, and Jie Tang. 2024{\natexlab{b}}.
\newblock Rest-mcts*: Llm self-training via process reward guided tree search.
\newblock \emph{arXiv preprint arXiv:2406.03816}.

\bibitem[{Zhang et~al.(2023)Zhang, Li, and Zou}]{Zhang2023ANT}
Ruoyu Zhang, Yanzeng Li, and Lei Zou. 2023.
\newblock A novel table-to-graph generation approach for document-level joint entity and relation extraction.
\newblock In \emph{Annual Meeting of the Association for Computational Linguistics}.

\bibitem[{Zhao et~al.(2023)Zhao, Zhou, Li, Tang, Wang, Hou, Min, Zhang, Zhang, Dong, Du, Yang, Chen, Chen, Jiang, Ren, Li, Tang, Liu, Liu, Nie, and rong Wen}]{Zhao2023ASO}
Wayne~Xin Zhao, Kun Zhou, Junyi Li, Tianyi Tang, Xiaolei Wang, Yupeng Hou, Yingqian Min, Beichen Zhang, Junjie Zhang, Zican Dong, Yifan Du, Chen Yang, Yushuo Chen, Z.~Chen, Jinhao Jiang, Ruiyang Ren, Yifan Li, Xinyu Tang, Zikang Liu, Peiyu Liu, Jianyun Nie, and Ji~rong Wen. 2023.
\newblock A survey of large language models.
\newblock \emph{ArXiv}, abs/2303.18223.

\bibitem[{Zheng et~al.(2023)Zheng, Wang, and Huang}]{Zheng2023ASO}
Hanwen Zheng, Sijia Wang, and Lifu Huang. 2023.
\newblock A survey of document-level information extraction.
\newblock \emph{ArXiv}, abs/2309.13249.

\bibitem[{Zhou et~al.(2023)Zhou, Li, Xiao, Yang, and Zhang}]{zhou2023llm}
Huixue Zhou, Mingchen Li, Yongkang Xiao, Han Yang, and Rui Zhang. 2023.
\newblock Llm instruction-example adaptive prompting (leap) framework for clinical relation extraction.
\newblock \emph{medRxiv}, pages 2023--12.

\bibitem[{Zhou et~al.(2024)Zhou, Zhang, Wang, Geng, Dong, and Tang}]{zhou2024lgb}
Ming Zhou, Dan Zhang, Yuandong Wang, Yangli-ao Geng, Yuxiao Dong, and Jie Tang. 2024.
\newblock Lgb: Language model and graph neural network-driven social bot detection.
\newblock \emph{arXiv preprint arXiv:2406.08762}.

\end{thebibliography}
\bibliographystyle{acl_natbib}

\appendix

\begin{table*}[t!]
\centering
\begin{tabular}{p{1.5cm}p{12cm}}
\hline
\textbf{Paradigms} & \textbf{Prompt} \\
\hline
\multirow{18}{*}{$\texttt{D-R-F}$} & Given a passage: \{sentences\}, and relation list: \{relation\_list\}\newline
                 Check the passage, and find which relations can be derived from the passage.\newline
                 Your output format is as following:\newline
                 relation1\newline
                 relation2\newline
                 ...\newline
                 one example like:\newline
                 country of citizenship\newline
                 father\newline
                 The relations must be in the relation list.\newline
                 If no relation in the sentence, you should  only output:\newline
                 no relation \\\cmidrule(l){2-2}
                 
& Given a relation: \{relation\}.\newline Provided a passage: \{sentences\}.\newline Derive all the triplet facts from the passage according to the given relations.\newline Your output format is as following:\newline [subject,\{relation\}, object]\newline [subject,\{relation\}, object]\newline ... \newline The subject and object should be an entity from the passage. 
\\\midrule
\multirow{26}{*}{\texttt{D-R-H-F}}
& Given a passage: \{sentences\}, and relation list: \{relation\_list\}\newline
                 Check the passage, and find which relations can be derived from the passage.\newline
                 Your output format is as following:\newline
                 relation1\newline
                 relation2\newline
                 ...\newline
                 one example like:\newline
                 country of citizenship\newline
                 father\newline
                 The relations must be in the relation list.\newline
                 If no relation in the sentence, you should  only output:\newline
                 no relation \\\cmidrule(l){2-2} 
& Given the relation: \{relation\}.\newline
Now the passage is: \{sentences\}.\newline
Derive all the entities from the passage that can serve as the subject of the \{relation\}.\newline
Your output format is as following:\newline
entity1\newline
entity2\newline
...\newline
The entities should all be from the passage.\\\cmidrule(l){2-2}

& Given the relation: \{relation\}.\newline
Now the passage is: \{sentences\}.\newline
Derive all the triplet facts from the passage that takes \{subject\} as a subject. \newline
Your output format is as following:\newline
[\{subject\},\{relation\},object]\newline
[\{subject\},\{relation\},object]\newline
...\newline
The object should be an entity from the passage.\\
\hline
\end{tabular}
\caption{ChatGPT prompt template for RE on Re-DocRED.}
\label{tab:chat-template}
\end{table*}

\label{sec:relation description}
\onecolumn
\begin{table}
\centering
\begin{tabular}{p{4cm}p{10cm}}
\hline
\textbf{Relation} & \textbf{Description} \\
\hline
\texttt{located in the administrative territorial entity} & In the ``located in the administrative territorial entity'' relation, the subject, a place, event, or item, resides or takes place in the object, an administrative region. Example: (Harvard University, located in the administrative territorial entity, Cambridge, Massachusetts).\\
\hline
\texttt{country} & For the ``country'' relation, the subject pertains to a non-human entity, such as an organization, place, or event. The object signifies the sovereign state where the subject is based or occurs. Example: (Amazon Inc, country, United States).\\
\hline
\texttt{country of citizenship} & The ``country of citizenship'' relation denotes that the subject, an individual, is recognized as a citizen by the object, a country. Example: (Elon Musk, country of citizenship, United States).\\
\hline
\texttt{contains administrative territorial entity} & The relation ``contains administrative territorial entity'' involves a subject, an administrative territory, encompassing the object, a subdivision or part of this administrative territory. Example: (California, contains administrative territorial entity, Los Angeles).\\
\hline
\texttt{has part} & The ``has part'' relation reflects that the subject, an entity or whole, comprises the object, a part or component of the subject. Example: (A car, has part, engine).\\
\hline
\texttt{date of birth} & In the ``date of birth'' relation, the subject, a person, was born on the object, the specified date. Example: (John Doe, date of birth, January 1, 1990).\\
\hline
\texttt{part of} & In the ``part of'' relation, the subject, a component or section, belongs to the object, a larger whole or aggregate. Example: (Engine, part of, a car).\\
\hline
\texttt{notable work} & The ``notable work'' relation indicates a significant work assigned to the subject, a creator, while the object is that noted scientific, artistic, or literary work itself. Example: (Jane Austen, notable work, Pride and Prejudice).\\
\hline
\texttt{publication date} & The ``publication date'' relation marks when the subject, a work, was first published or released, with the object being that specific date. Example: (Pride and Prejudice, publication date, 1813).\\
\hline
\texttt{inception} & In the ``inception'' relation, the subject, an event, or an item (not a person), came into existence at the object, a specific date or point in time. Example: (Google, inception, September 4, 1998).\\
\hline
\texttt{date of death} & The ``date of death'' relation specifies when the subject, a once-living person, died. The object is the particular date of demise. Example: (Albert Einstein, date of death, April 18, 1955).\\
\hline
\hline
\end{tabular}
\caption{New designed relation descriptions. We only present part of the descriptions of 96 relations. The whole relation descriptions can be found via this link: \url{https://github.com/THUDM/AutoRE}.
}
\label{tab:Relation Description}
\end{table}

\begin{table*}[t!]
\centering
\begin{tabular}{p{4cm}p{10cm}}
\hline
\textbf{Submission} & \textbf{Instruct Tuning Template} \\
\hline
\texttt{relation\_template} & Given a passage: \{sentences\}, list any underlying relations. \\
\hline
\texttt{entity\_template}  & Given a relation \{relation\}, and its description: \{description\} and a passage: \{sentences\}, list entities that can be identified as suitable subjects for the relation. \\
\hline
\texttt{fact\_template}  & Given relation \{relation\} and relation description: \{description\}. Provided a passage: \{sentences\}, list all triple facts that take \{relation\} as the relation and \{subject\} as the subject.  \\
\hline
\end{tabular}
\caption{Instruction tuning template for RHF.}
\label{tab:instructtuning-template}
\end{table*}
\label{sec:relation result}

\end{document}